\documentclass[sigconf,nonacm]{aamas} 

\usepackage{balance} 

\newtheorem{example}{Example}[section]
\newcommand{\methodName}{\texttt{PADME}}
\newcommand{\methodNameFull}{\texttt{P}rocedure \texttt{A}ware \texttt{D}yna\texttt{M}ic \texttt{E}xecution}
\newcommand{\methodNameFullNoFormat}{Procedure Aware DynaMic Execution}
\usepackage{algorithm}
\usepackage{comment}
\usepackage{algpseudocode}
\usepackage{tikz}
\usetikzlibrary{trees}
\usepackage{array}
\usepackage{booktabs}
\usepackage{amsmath} 
\usepackage{listings}
\usepackage{pgfplots}
\pgfplotsset{compat=1.9}
\usepackage{subcaption}
\usepackage{xcolor}

\definecolor{colAct}   {RGB}{230,97,1}    
\definecolor{colReAct} {RGB}{158,1,66}  
\definecolor{colCoT}   {RGB}{255,195,0}  
\definecolor{colSPR}   {RGB}{0,158,115}    
\definecolor{colOurs}  {RGB}{0,114,178} 

\title{\methodName{}: \methodNameFullNoFormat{}}

\author{Deepeka Garg}
\affiliation{
  \institution{J.P. Morgan AI Research}
\country{United Kingdom}}
\email{deepeka.garg@jpmorgan.com}

\author{Sihan Zeng}
\affiliation{
  \institution{J.P. Morgan AI Research}
\country{USA}}
\email{sihan.zeng@jpmchase.com}

\author{Annapoorani L. Narayanan}
\affiliation{
  \institution{J.P. Morgan AI Research}
\country{United Kingdom}}
\email{annapoorani.lakshminarayanan@jpmorgan.com}

\author{Sumitra Ganesh}
\affiliation{
  \institution{J.P. Morgan AI Research}
\country{USA}}
\email{sumitra.ganesh@jpmorgan.com}

\author{Leo Ardon}
\affiliation{
  \institution{J.P. Morgan AI Research}
\country{United Kingdom}}
\email{leo.ardon@jpmorgan.com}

\begin{abstract}
Learning to autonomously execute long-horizon procedures from natural language remains a core challenge for intelligent agents. Free-form instructions such as recipes, scientific protocols, or business workflows encode rich procedural knowledge, but their variability and lack of structure cause agents driven by large language models (LLMs) to drift or fail during execution. We introduce \methodNameFull{} (\methodName{}), an agent framework that produces and exploits a graph-based representation of procedures. Unlike prior work that relies on manual graph construction or unstructured reasoning, \methodName{} autonomously transforms procedural text into executable graphs that capture task dependencies, decision points, and reusable subroutines. Central to \methodName{} is a two-phase methodology; \textit{Teach} phase, which focuses on systematic structuring, enrichment with executable logic of procedures, followed by \textit{Execute} phase, which enables dynamic execution in response to real-time inputs and environment feedback. This separation ensures quality assurance and scalability, allowing expert knowledge to be encoded once and reliably reused across varying contexts. The graph representation also provides an inductive bias that reduces error accumulation in long-horizon reasoning, underscoring the importance of structured procedure modeling for reliable agent-driven automation. Empirically, \methodName{} achieves state-of-the-art performance on four diverse benchmarks, including ALFWorld and ScienceWorld. These results demonstrate that agents equipped with graph-based procedure representations offer a powerful intermediate abstraction for robust and generalizable execution.

\end{abstract}

\keywords{Autonomous Agents, LLMs, Graph Representation, Workflow Automation.}

\newcommand{\BibTeX}{\rm B\kern-.05em{\sc i\kern-.025em b}\kern-.08em\TeX}

\setcopyright{none}
\begin{document}


\pagestyle{fancy}
\fancyhead{}


\maketitle 


\section{Introduction}
\label{sec.introduction}
\begin{figure}[!t]
    \centering
    \includegraphics[width=0.45\textwidth]{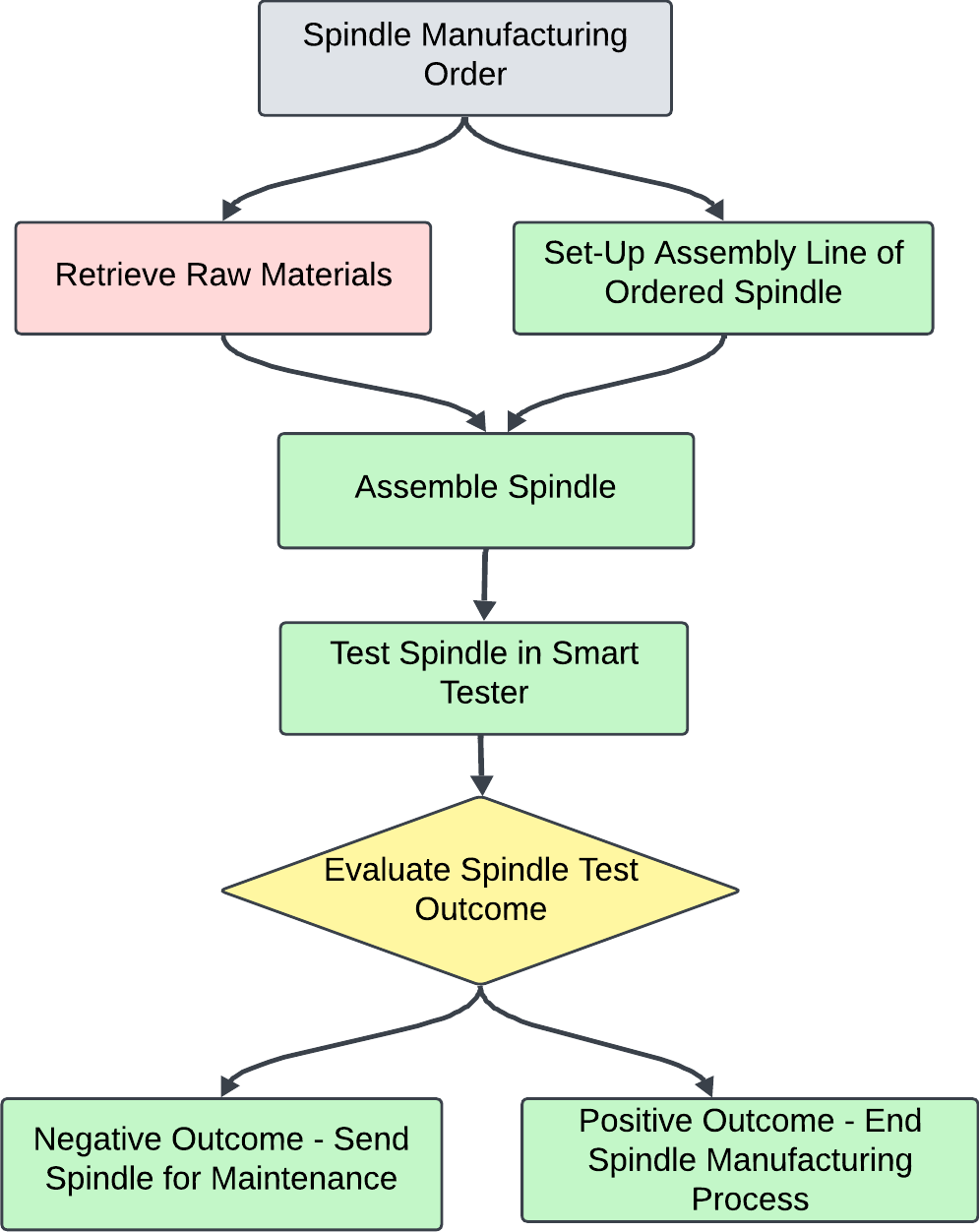}
    \caption{Decision Graph Example}
    \label{fig:dagexample}
\end{figure}
Real-world tasks and procedures often involve long horizons, conditional dependencies, and diverse action spaces, causing current large language model (LLM) agents to drift, accumulate errors, or fail to maintain coherent execution. Tasks such as following a recipe, conducting a scientific experiment, or completing a business workflow differ widely in domain, action space, and temporal structure. While LLM agents can reliably follow short, single-goal prompts, they often lose coherence when reasoning over diverse domains with dependencies that span dozens of steps \cite{fujisawa2024procbench}. This limitation highlights the need for methodologies that can generalize across tasks while maintaining consistency over extended horizons. Much of the knowledge required to perform such tasks is written in free-form text stored in standard operating procedures (SOPs), manuals, incident logs, and training documents designed for humans rather than agents. Rewriting these resources into formal representations is feasible in theory, but impractical at scale, motivating the automated learning of structured intermediate formats that bridge free form text and executable procedures. Prior work shows that free-form manuals can guide goal-directed AI systems \cite{tessler2021learning}, but their lack of structure, inconsistent syntax, and human-oriented design continues to make robust execution challenging.


\begin{figure*}[!t]
  \centering
  \includegraphics[width=0.9\textwidth]{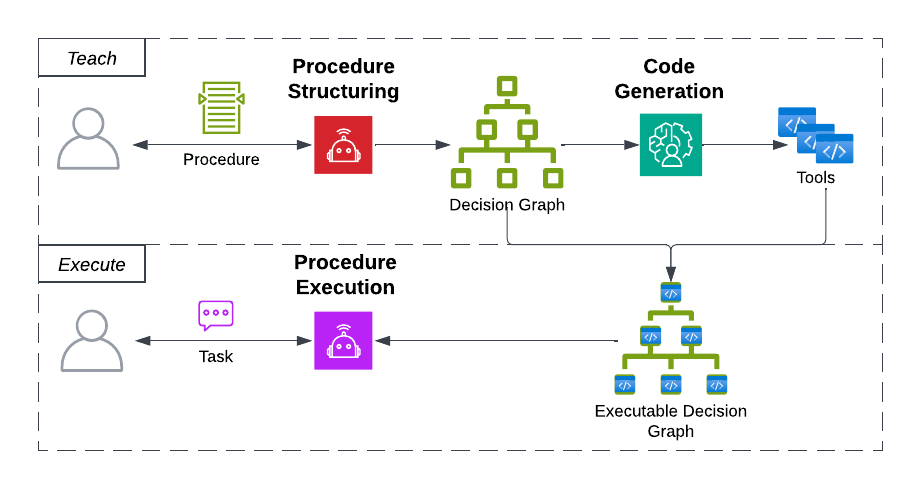}
  \caption{\methodName{}: \textit{Teach} and \textit{Execute} Framework}
  \label{fig:methodology}
\end{figure*}

To address this problem, we introduce \methodNameFull{} (\methodName), a two-phase agent framework. In the \textit{Teach} phase, a structuring agent converts free-form procedures into executable decision graphs; in the \textit{Execute} phase, an execution agent follows the graph in real time, adapting the path to live observations. The decision graph is the core representation of the procedures, where nodes correspond to steps and edges encode logical or temporal dependencies. We call this representation a decision graph because procedures typically include branching choices, which our structuring process captures explicitly as decision nodes, enabling conditional reasoning. For example, in one of our evaluation datasets \cite{monti2024nl2processops}, a spindle manufacturing procedure (Figure~\ref{fig:dagexample}) begins with arranging raw materials and setting up the assembly line. After the spindle is assembled and tested in a smart tester, the graph reaches a decision node (\textit{Evaluate Spindle Test Outcome}). If the outcome is negative, the spindle is sent for maintenance and if positive, the process concludes successfully. These graphs are both human-readable for domain experts and machine-readable for agents, providing a verifiable and automatable blueprint for procedure execution.





Recent work has shown that graph-based representations can structure LLM reasoning for reliable task execution, but most approaches rely on manual graph construction. AgentKit \cite{wu2024agentkit} requires users to break a task into LEGO-like pieces and hand-build a dependency graph that controls the LLM’s output. SPRING \cite{wu2024spring} parses \LaTeX\ game descriptions into a context string and users assemble a chain-of-thought graph whose final node becomes an action. In contrast, \methodName{} automatically constructs decision graphs from free-form text, eliminating the need for manual engineering. Unlike methods tied to fixed simulators or action spaces \cite{min2021film, huang2022language}, \methodName{} treats procedural text itself as the primary knowledge source. In the \emph{Teach} phase, the procedure structuring agent builds a semantic decision graph independent of tools (execution logic), with tool bindings attached to individual nodes at a later stage. By modularizing semantic structuring from tool binding, \methodName{} preserves a stable logical representation that can adapt to new environments or tool sets without requiring procedures to be restructured.

Our key contribution is a modular two-phase approach that separates \textit{Teach} (automatic conversion of free-form procedures into an executable \textit{decision graph} with dependencies, decision points, and node-level inputs/outputs, pre/post conditions metadata) from \textit{Execute} (context-aware decision graph execution with tool invocation). Figure~\ref{fig:methodology} illustrates the full pipeline. We evaluate \methodName{} across four domains spanning office workflows, cooking, laboratory science, and household manipulation. Evaluation benchmarks include a business-process corpus executed via APIs, a consolidated recipe book requiring dynamic dish selection, and the long-horizon simulators ScienceWorld and ALFWorld. The same \textit{Teach} and \textit{Execute} phases are applied unmodified across all benchmarks, demonstrating domain-agnostic generalization across diverse tasks and action vocabularies.
\begin{figure*}
    \centering
    \includegraphics[width=0.95\textwidth] {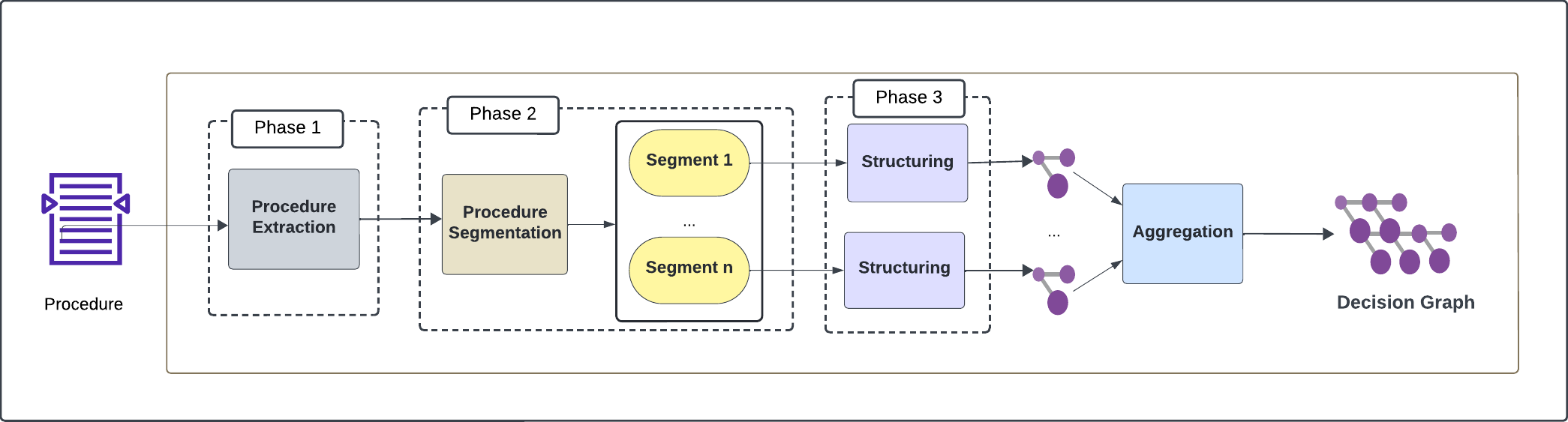}
    \caption{Procedure Structuring Agent}
    \label{fig:structuringmethodology}
\end{figure*}

\section{\methodName{} Methodology} 
\label{sec.methodology}
\methodName{} transforms unstructured procedural text into a structured decision graph, which is then \emph{executed} with real-time, context-aware reasoning and decision making. The end-to-end methodology consists of two phases; \textit{Teach} and \textit{Execute} as illustrated in Figure~\ref{fig:methodology}.


\subsection{Theoretical Foundations of \methodName{}}
\label{sec.theorecticaldetails}
We now formalize the decision graph representation used in \methodName{} and analyze its implications for error control.
\subsubsection{Decision Graph Representation}

A \emph{decision graph} is a directed acyclic graph $G = (V, E)$ where each node $v \in V$ is a typed operator with signature
\[
f_v : \mathcal{X}_v \rightarrow \mathcal{Y}_v,
\]
where $\mathcal{X}_v$ and $\mathcal{Y}_v$ are input and output spaces, respectively. Edges $(u,v) \in E$ encode the dependency that the outputs of $f_u$ may be required as inputs to $f_v$.

We categorize a node $v$ into the following five classes,
\[
C(v) \in 
\begin{aligned}[t]
\{&\textsc{HumanInput}, \textsc{InfoProcessing}, \\
  &\textsc{InfoExtraction}, \textsc{Knowledge}, \textsc{Decision}\},
\end{aligned}
\]
which represent user-provided data, internal computation, external retrieval, reference information, 
and conditional branching, respectively (we explain the categories in detail in Sec.~\ref{sec.procedurestructuring}). In particular, \textsc{Decision} nodes are parameterized as conditional distributions
\[
p(\text{branch} \mid \text{context}(h)),
\]
where $\text{context}(h)$ summarizes the execution history up to node $v$. Thus, decision graphs unify deterministic procedural dependencies with stochastic or context-sensitive branching.

\begin{figure*}[t]
    \centering
    \includegraphics[width=0.85\textwidth,height=6.5cm]{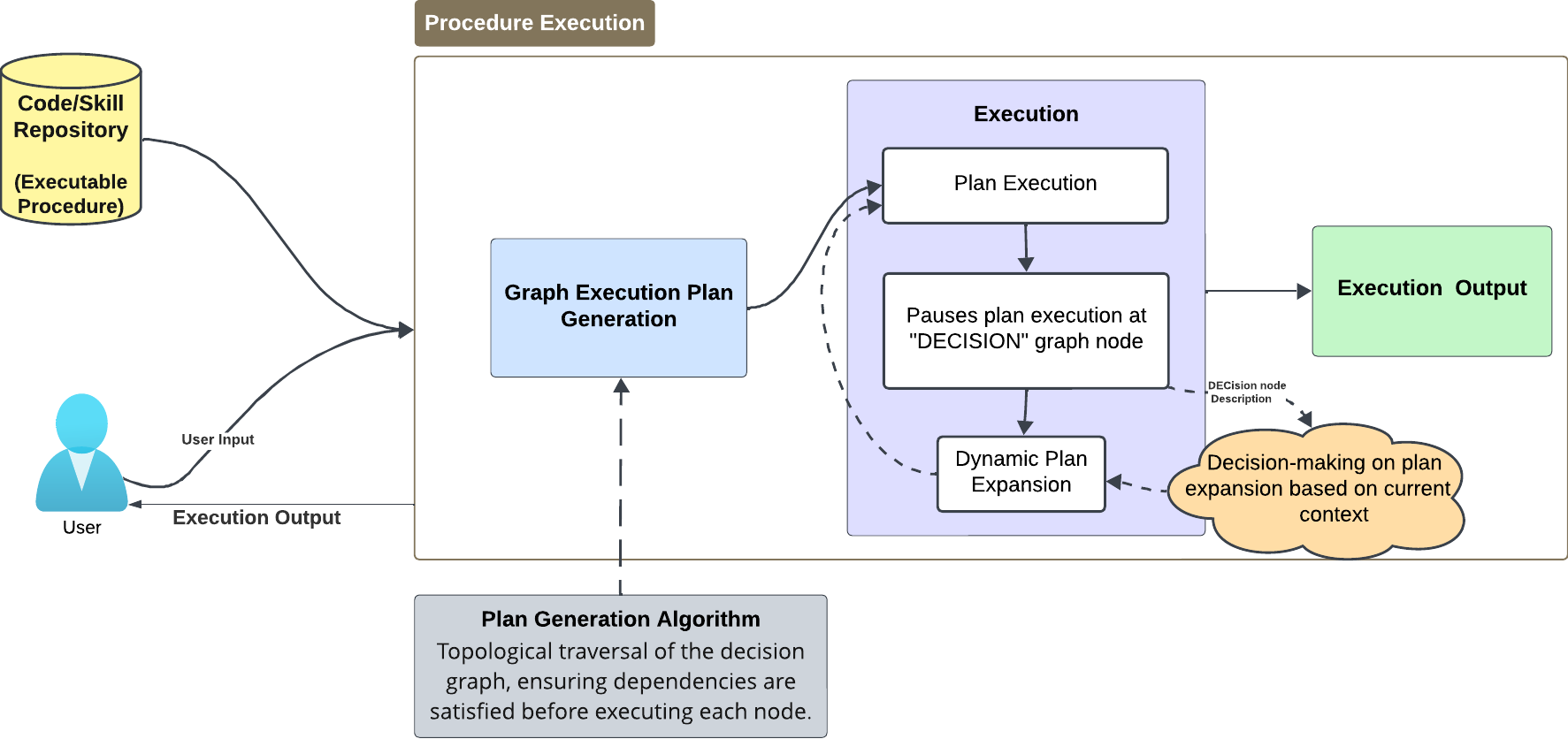}
    \caption{Procedure Execution Agent}
    \label{fig:executionmethodology}
\end{figure*}

\subsubsection{Complexity and Error Decomposition under Teach--Execute Separation}

During \textit{Teach}, free-form text is mapped into a decision graph, resolving structural ambiguity in advance and enforcing \emph{topological consistency}. During \textit{Execute}, reasoning is restricted to resolving operators $f_v$ and \textsc{Decision} category nodes, constraining the action space to valid paths within $G$ and preventing drift into irrelevant or incoherent steps. 

Executing procedures directly from free-form text requires reasoning over procedure step sequences of length $T$ with branching factor $|\mathcal{A}|$, yielding a search space of size $\Theta(|\mathcal{A}|^T)$. LLMs approximate this process autoregressively, but remain vulnerable to drift when $T$ is large. In contrast, decision graphs constrain execution to valid paths defined by $G=(V,E)$. 
Since each step in the original procedure becomes a node, we have $|V| = O(T)$ and 
$|E| = O(T)$ in linear chains, or $|E| = O(T^2)$ in dense cases where every step 
depends on many others. Thus, the worst-case traversal cost is $O(|V|+|E|) = O(T^2)$, 
independent of the branching factor $|\mathcal{A}|$. 
This contrasts with unstructured free-form execution, which must search over 
$O(|\mathcal{A}|^T)$ possible sequences. 

Two types of errors can arise at runtime. 
\emph{Branching errors} occur at $B$ decision points ($B \ll T$), where $B$ is the number of decision nodes in the graph and $T$ the total number of steps in the free-form execution. 
At these nodes, the LLM must select among alternative paths. 
\emph{Parameterization errors} arise when providing inputs to node functions $f_v$. 
In unstructured free-form execution, this requires searching over a combinatorial space of size $\Theta(N \times P)$, where $N$ is the number of procedure action steps and $P$ the parameter space of actions. In contrast, decision graphs restrict admissible parameters for each node $v$ to the set $\mathcal{X}_v \subseteq \{\text{outputs}(u) : (u,v) \in E\} \cup \mathcal{I}$, where $\mathcal{I}$ denotes the initial task inputs. This reduces the parameter search space from potentially unbounded $P$ (e.g., free-form strings) to a bounded set derived from explicit graph dependencies, limiting error propagation.

This is in addition to one-time structuring errors from the \textit{Teach} phase, which can be corrected through human or automated validation (see supplementary material for details) and therefore do not accumulate during execution. The accumulation of run-time errors is dominated by decision nodes and localized parameterization, rather than the full horizon $T$.

\subsection{\textit{Teach} Phase: Making Procedures AI‑Ready}
In the \textit{Teach} phase, raw procedural text whether a written SOP, a set of instructions, or a troubleshooting guide is converted into a \emph{decision graph} that encodes the procedure’s logic. In this graph, nodes represent individual steps (including decision points), while edges capture logical flow, conditions, and temporal dependencies. For critical domains such as finance or healthcare, the decision graph can be validated by human domain experts in addition to our automated evaluation module (see details in supplementary material), ensuring correctness and completeness. Once validated, each node is enriched with executable bindings such as code snippets or API calls. This enrichment may be performed by human engineers or delegated to LLMs, which our framework supports given their demonstrated capabilities in code generation. 
By the end of \textit{Teach} phase, the originally free-form procedure is converted into an AI-ready executable decision graph. 

\begin{algorithm}[t]
\caption{Graph Traversal Algorithm of \methodName{}}\label{alg:grap_traversal}
\begin{algorithmic}[1]
\State $G$: decision graph
\State $S$: set of initial nodes to start traversing the graph from
\State $P$: ordered list containing the nodes already executed
\State
\State $Q \gets$ empty queue that will contain the nodes to analyze
\State $V \gets$ empty set that will contain the nodes visited
\State
\State $Q \gets S; V \gets S$

\While{$Q$ is not empty}
    \State $n \gets Q$.pop()
    \State
    \If{$\{\exists p \in G.\textsc{parents}(n);\; G.\textsc{isDecision}(p)\}$}
        \State\hspace{-\algorithmicindent}{\color{teal} /* $n$ depends on a DECISION node. Stop the exploration of this branch of the graph as we need to wait for the decision to be made to proceed */}
        \State continue
    \EndIf
    \State
    \If{$(n \in S)$ $\vee$ $(\forall d \in G.\textsc{parents}(n);\; d \in P)$}
        \State\hspace{-\algorithmicindent}{\color{teal} /* $n$ has all its dependencies planned for execution we can add it to $P$ */}
        \State $P \gets P + [n];$
    \Else \If {$(G.\textsc{parents}(n) \cup Q \neq \emptyset)$}
        \State\hspace{-\algorithmicindent}{\color{teal} /* $n$ has at least one dependency yet to be analyzed and therefore not planned for execution, add it back to $Q$ */}
        \State $Q \gets Q + [n];$
    \EndIf
    \EndIf
    
    \State

    \ForAll{$c \in G.\textsc{children}(n)$}
        \If {$c \notin V$}
            \State $Q \gets Q + [c]; V \gets V + [c];$
        \EndIf
    \EndFor
\EndWhile
\State \Return $P$
\end{algorithmic}
\end{algorithm}

\subsubsection{Procedure Structuring (Decision Graph Creation) Agent}
\label{sec.procedurestructuring}
The Procedure Structuring (Figure~\ref{fig:structuringmethodology}) is an LLM-based agent that interprets and organizes procedural text into a decision graph. It first filters out non-actionable content such as administrative headers, legal disclaimers, or commentary, retaining only actionable steps. The remaining procedure is segmented into smaller units ${S_k}_{k=1}^m$; numbered instructions are split deterministically so that each step becomes a segment, while free-form prose is segmented at topic shifts detected by the structuring agent. This segmentation prevents long passages from obscuring fine-grained details and ensures that local dependencies are preserved.

Each segment is then transformed into a local decision subgraph, a Directed Acyclic Graph (DAG) that captures the logic of that part of the procedure. The global graph built so far is always provided as context for subsequent segments, ensuring that cross-segment dependencies are preserved. Once all segments are processed, the subgraphs are merged into a single DAG $G = (V, E)$, where nodes correspond to subtasks or decisions and edges encode dependencies and execution flow. Nodes are annotated with metadata following a user-defined schema that specifies the name, description, input/output fields, dependencies, and category. Categories help define node operation and fall into five types:
\begin{itemize}
\item \textbf{Human Input}: Requires user-provided data.
\item \textbf{Information Processing}: Performs internal computation or transformation.
\item \textbf{Information Extraction}: Pulls in information from external sources such as APIs.
\item \textbf{Knowledge}: Provides reference or background material.
\item \textbf{Decision}: Encodes branching based on conditional logic.
\end{itemize}

These categories cover all operations observed across business, cooking, household, and science procedures, however, additional categories can be added when needed.

Figure~\ref{fig:dagexample} illustrates an example decision graph from the spindle manufacturing procedure. A \emph{Human Input} node (grey) initiates the process with a spindle order, which triggers an \emph{Information Extraction} node (pink) to retrieve raw materials and an \emph{Information Processing} node (green) to set up the assembly line. The flow then continues through additional \emph{Information Processing} nodes that assemble and test the spindle. At this point, a \emph{Decision} node (yellow) evaluates the spindle test outcome, branching either to maintenance in the case of a negative result or to successful completion if the outcome is positive. Because the procedure is visualized as a graph, domain experts can review, spot errors, and edit nodes or edges with far less cognitive load than reading raw text.

To make the graph executable, each node except those categorized as \emph{Decision} is equipped with an executable function. \emph{Decision} nodes are not assigned executable functions, instead, during execution, the execution agent selects branches dynamically based on context such as upstream outputs, environment state, or user input. Returning to the spindle manufacturing example, the test outcome is evaluated at runtime, sending the process to maintenance if the spindle fails testing or to successful completion if it passes.

\subsection{\textit{Execute} Phase: Dynamic and Reusable Execution of AI-ready Procedures}
\label{sec.executionphase}
After the \emph{Teach} phase produces an AI-ready decision graph, the \emph{Execute} phase reuses this representation across diverse task instances. New tasks with varying input parameters are handled by the same underlying graph, which adapts dynamically to both inputs and live conditions. This design enables adaptive execution, ensuring consistency while allowing flexibility. The procedure execution agent is described below:

\subsubsection{Procedure Execution Agent}
\label{sec.execagent}
The execution process is facilitated by decision graph, akin to the structured representation of a plan. This graph serves as a blueprint that delineates the sequence of actions to execute to accomplish a goal, thereby providing a systematic approach to guide the LLM agent execution and minimizing ambiguity in the execution pathway. Nevertheless, a degree of dynamism is crucial to solve real-life scenarios, as processes frequently necessitate decisions that are contingent upon the context encountered during execution. To accommodate this requirement, we introduce the concept of decision nodes within the decision graph, leveraging the categorization of the nodes done by our procedure structuring agent. As stated previously, these nodes are designed to enable context-sensitive decisions at execution time. 

We formalize execution as a graph traversal problem (Algorithm~\ref{alg:grap_traversal}). The algorithm proceeds in topological order through the decision graph, invoking executable nodes sequentially. When a decision node is encountered, traversal pauses and the execution agent uses the current context (including upstream outputs, environment state, or user input) to select the appropriate branch. The graph is then expanded along the chosen edge, and traversal continues. This process repeats until no further nodes remain to be executed. This design frames procedure execution as a form of structured reasoning; deterministic where the graph specifies explicit dependencies, but adaptive where decision nodes defer choice to runtime. As a result, the traversal algorithm reduces error accumulation, while still allowing flexibility to adapt execution paths to live conditions.

\section{Experiments}
\label{sec.experiments}
Our experiments have been conducted using GPT-4, chosen for its strong performance across a variety of tasks, including solving chess puzzles, math problems, and coding challenges \cite{bubeck2023sparks}. However, \methodName{} is model-agnostic and GPT-4 can be replaced with other language models. 

\subsection{Datasets and Benchmarks}
Open-source SOP corpora are scarce because most LLM benchmarks contain isolated tasks such as solving math problems \cite{zeng2023mr} or generating code \cite{peng2024humaneval}. Benchmarks that explicitly target multi-step instruction following (aka procedures), typically focus on toy procedural instructions (e.g., string manipulation coding tasks) rather than realistic procedures \cite{fujisawa2024procbench}. To address this gap, we evaluate \methodName{} on four datasets where full procedures already exist or can be reliably inferred, spanning household and science environments, recipes, and real-world business processes.


\begin{itemize}
     \item \textbf{Business Process Dataset} \cite{monti2024nl2processops}: 
     A collection of business procedures including linear, branched, and concurrent flows. For each procedure, we create three task variants; \emph{easy} (straightforward allowing direct execution of steps), \emph{medium} (with irrelevant details/noise requiring filtering), and \emph{hard} (with ambiguity requiring additional decision making). For example, from the spindle manufacturing SOP: “How is the spindle manufacturing process executed at HSD company?” (\emph{easy}); “I want to either get a spindle manufactured or a gear manufactured, for whichever process is available. How do I proceed?” (\emph{medium}, added an additional competing process); “The spindle manufacturing process is about to start. How will execution proceed if the smart tester outcome is negative? Can you do the execution for me?” (\emph{hard}, ambiguous state).

    \item \textbf{Recipe Dataset}: Recipes resemble business SOPs as multi-step, conditional workflows dependent on available inputs (ingredients). Our dataset extends RecipeNLG \cite{bien2020recipenlg} by aggregating individual recipes into a unified “recipe book” akin to a chef’s SOP manual. In the \textit{Teach} phase, the book is converted into a single decision graph with $\sim$80 nodes capturing the logic of all dishes. At execution time, the \textit{Execute} phase interprets a task query, selects the appropriate recipe, and performs the steps while handling optional ingredients or substitutions. Task queries include three variants; \emph{easy} (direct ingredient-to-recipe mapping), \emph{medium} (noise from extra, irrelevant ingredients) and \emph{hard} (ambiguous queries such as “What can I make with these ingredients?”).

    \item \textbf{ScienceWorld} \cite{wang2022scienceworld}: An interactive environment for evaluating scientific reasoning, simulating experiments of 20–50 steps such as gathering equipment, preparing materials, and recording measurements. Originally developed for reinforcement learning, we repurpose ScienceWorld as a procedure execution benchmark by converting expert demonstrations into SOPs and executing them via its text interface. Script generation requires real-time decisions based on environmental feedback. Notably, task-completion trajectories in ScienceWorld are not unique. To compute evaluation scores, we extract and compare key checkpoints against ground truth trajectories. As an RL environment, ScienceWorld rewards each step towards the goal, with rewards being positive only at significant milestones. Key checkpoints are those generating positive rewards. The final match is a $0$/$1$ score indicating successful task completion.
    
    \item \textbf{ALFWorld} \cite{shridhar2021alfworld}: ALFWorld is similar in nature to ScienceWorld, originally designed to be a testbed for language-based reinforcement learning methods. Tasks in ALFWorld involve locating objects and performing operations on them such as moving, cleaning, heating, etc., typically requiring 5–15 steps to complete. We repurpose ALFWorld for procedure execution by converting expert demonstrations into SOPs.
\end{itemize}

Each dataset is paired with a domain-specific action library to enable end-to-end execution for \methodName{} and all baselines. The Business Process dataset provides Python APIs while for the Recipe dataset we implemented primitive cooking actions (e.g., \texttt{bake}, \texttt{stir}). In ALFWorld and ScienceWorld, the tool libraries are the collections of valid actions that can be executed in the respective environments. Ground-truth execution traces are defined as the sequence of tool calls that completes each procedure, with success in ALFWorld and ScienceWorld determined by whether the resulting state transitions match the target outcomes.

\begin{figure*}[t]
\centering
\begin{tikzpicture}
\begin{axis}[
    ybar,
    bar width=15pt,
    width=1\textwidth,
    height=5.5cm,
    enlarge x limits=0.2,
    ylabel={Average FM Score},
    symbolic x coords=
    {Recipe Dataset, ALFWorld, ScienceWorld, Business Process Dataset},
    xtick=data,
    ymin=0,ymax=1,
    legend style={font=\scriptsize,at={(0.5,-0.3)},anchor=north,legend columns=5},
    area legend,
    nodes near coords,
    nodes near coords align={vertical},
    every node near coord/.append style={font=\tiny, text=black},
]
\addplot+[fill=green!40, draw=green!40] coordinates {(Recipe Dataset,0.28) (ALFWorld,0.62)(ScienceWorld,0.62) (Business Process Dataset,0.71)};
\addplot+[fill=red!60, draw=red!60] coordinates {(Recipe Dataset,0.31) (ALFWorld,0.58)(ScienceWorld,0.44) (Business Process Dataset,0.60)};
\addplot+[fill=brown!80, draw=brown!80] coordinates {(Recipe Dataset,0.24) (ALFWorld,0.33)(ScienceWorld,0.63) (Business Process Dataset,0.74)};
\addplot+[fill=black, draw=black]  coordinates {(Recipe Dataset,0.21) (ALFWorld,0.58)(ScienceWorld,0.22) (Business Process Dataset,0.65)};
\addplot+[fill=blue, draw=blue] coordinates {(Recipe Dataset,0.87) (ALFWorld,0.69)(ScienceWorld,0.74) (Business Process Dataset,0.74)};
\legend{ACT-Only,ReAct,C-o-T,SPRING,\methodName{} (ours)}
\end{axis}
\end{tikzpicture}
\caption{Final Match (FM) scores averaged over five independent trials on each of the four datasets (higher is better).}
\label{fig:aggregate-FM}
\end{figure*}
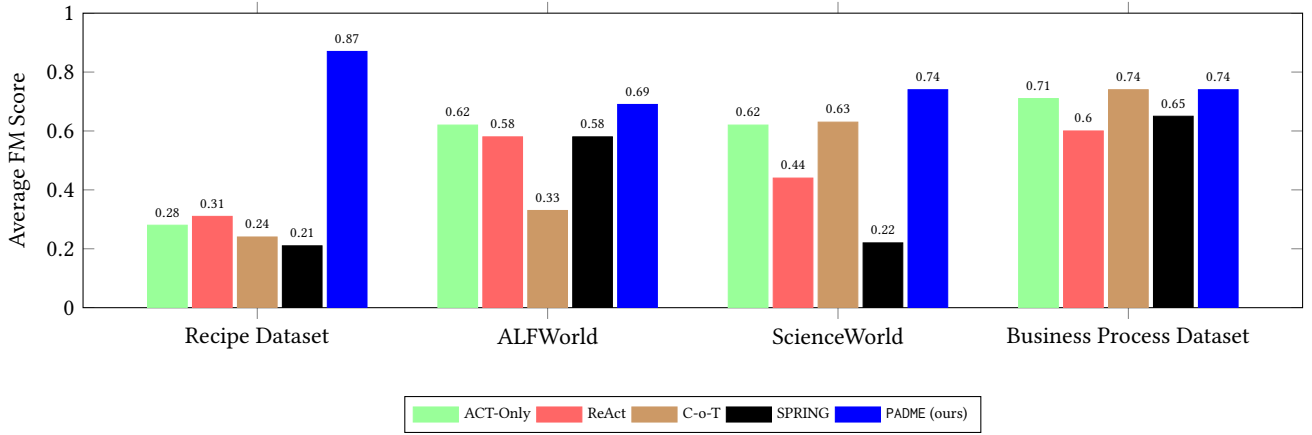

\subsection{Evaluation Metrics}
We evaluate \methodName{} and baselines by comparing their predicted action sequences against ground-truth action trajectories. For \methodName{}, this end-to-end evaluation reflects both decision graph quality and execution performance, since any structural error in the graph will lead to execution failure. While \methodName{} includes a separate pipeline for analyzing decision graph quality (see supplementary material for details), we report only end-to-end results here for fairness, as some baselines do not include an explicit structuring stage. Four metrics are reported \cite{fujisawa2024procbench}: (i) Prefix Match Length (PML): longest correct prefix; (ii) Prefix Accuracy (PA): that prefix length normalized by sequence length; (iii) Sequential Match (SM): exact 0/1 match of the full sequence; and (iv) Final Match (FM): 0/1 correctness of the last action. PML and PA measure how long a method stays aligned before drifting, while SM and FM capture end-to-end success. Together these provide a fine-grained view of partial vs. complete correctness. Full mathematical definitions of these metrics are presented below.

\subsubsection{\methodName{} Evaluation Metrics}
\label{appen.evaluationmetrics}
Let $T = (t_1, \dots, t_N)$ denote the target sequence of length $N$. Let $P = (p_1, p_2, \dots, p_M)$ denote the predicted sequence of length $M$. We report:

\begin{itemize}
\item \textbf{Prefix Match Length (PML):}
This metric measures the length of the longest prefix where the predicted and target sequences exactly match. Formally, let $k$ be the largest index such that $ t_i = p_i $ for all $ 1 \leq i \leq k $:
$$
\text{PML}(T, P) = \max \left( k \mid t_i = p_i \text{ for all } 1 \leq i \leq k \right)
$$

\item \textbf{Prefix Accuracy (PA):}  
Using the PML, Prefix Accuracy normalizes the match length by the longer of the two sequences:
\[
\text{PA}(T, P) = \frac{\text{PML}(T, P)}{\max(N, M)}
\]
This penalizes both over- and under-prediction, making it robust to output length variation.

\item \textbf{Sequential Match (SM):}  
A binary indicator of exact match across the full predicted and ground truth sequences:
\[
\text{SM}(T, P) =
\begin{cases}
    1 & \text{if } \text{PA}(T, P) = 1, \\
    0 & \text{otherwise}
\end{cases}
\]

\item \textbf{Final Match (FM):}  
A binary metric indicating whether the final predicted element matches the final target element, regardless intermediate discrepancies:
\[
\text{FM}(T, P) =
\begin{cases}
    1 & \text{if } t_N = p_M, \\
    0 & \text{otherwise}
\end{cases}
\]

\end{itemize}

\subsection{Baselines}
We compare \methodName{} against baseline methods representative of current approaches to natural language task planning and execution. For fairness, all baselines and \methodName{} use GPT-4, share the same tool library, and are evaluated on the same task queries. While baselines operate directly on unstructured procedure text, \methodName{} first converts it into a decision graph.

\begin{itemize}
\item \textbf{Act-Only} \cite{yao2023react}: Directly generates environment actions (e.g., navigation commands, API calls) from observations, with no explicit intermediate reasoning.   

\item \textbf{ReAct} \cite{yao2023react}:  Interleaves short reasoning statements with actions, enabling adaptive in-flight planning and interpretable decision paths.  

\item \textbf{Chain-of-Thought (CoT)} \cite{wei2022chain}: Produces a full natural language reasoning chain before outputting actions. Unlike ReAct, reasoning is offline and does not interact with the environment during inference.  

\item \textbf{SPRING (Studying the Paper and Reasoning to Play Games)} \cite{wu2024spring}: A zero-shot language agent that combines document understanding with structured reasoning to guide action selection. It first extracts knowledge from task documentation, then uses a manually crafted graph of interdependent questions to reason over the graph in topological order, and finally translates the final answer into an executable action.
\end{itemize}

Note that we did not include AgentKit \cite{wu2024agentkit} as a baseline, since it requires users to manually decompose tasks, construct node-level graphs, and craft prompts for each node. Such extensive engineering is impractical for large-scale evaluation. In contrast, \methodName{} generates the entire graph automatically from raw procedures, eliminating this bottleneck.

\section{Results and Analysis}
End-to-end execution results across all tasks are summarized over five independent runs for each dataset (Table \ref{tbl:evaluation-results}). Figure~\ref{fig:aggregate-FM} shows the FM scores. For \methodName{}, end-to-end results capture both the quality of graph structuring in the \emph{Teach} phase and execution in the \emph{Execute} phase, providing a fair comparison to baselines that operate directly from free-form procedures.

On the \textbf{Recipe} dataset, \methodName{} outperforms all four baselines with the average PML being $2.64\pm0.28$, more than doubling the best competitor (SPRING at $0.96\pm0.04$), while PA is $0.69\pm0.07$ versus $0.22\pm0.01$ for the strongest baseline. Recipes must be executed in the exact written order as one cannot bake before mixing, therefore, SM is an important metric. \methodName{}'s SM score of $0.64\pm0.08$ outerperforms the baselines' SM. Finally, \methodName{} achieves an FM score of $0.87\pm0.01$ compared with best baseline ($0.31\pm0.01$ for ReAct). 

\begin{table}[h]
\centering
\caption{Results are summarized over 5 independent trials. \methodName{} achieves consistently high performance across all datasets.}
\label{tbl:evaluation-results}
\begin{tabular}{@{}lcccc@{}}
\toprule
&\multicolumn{4}{c}{Metric}\\
\cmidrule(l){2-5}
Method&\multicolumn{1}{c}{PML}&\multicolumn{1}{c}{PA}&\multicolumn{1}{c}{SM}&\multicolumn{1}{c}{FM}\\
\midrule
\multicolumn{5}{c}{\bfseries Recipe Dataset}\\
\midrule
Act-Only            & $0.87\pm0.06$ & $0.22\pm0.01$ & $0.07\pm0.01$ & $0.28\pm0.01$\\
ReAct               & $0.75\pm0.01$ & $0.17\pm0.00$ & $0.05\pm0.00$ & $0.31\pm0.01$\\
C-o-T               & $0.62\pm0.03$ & $0.14\pm0.01$ & $0.03\pm0.01$ & $0.24\pm0.02$\\
SPRING              & $0.96\pm0.04$ & $0.21\pm0.01$ & $0.04\pm0.01$ & $0.21\pm0.03$\\
\methodName{} (ours)& \textbf{2.64 $\pm$0.28} & \textbf{0.69 $\pm$ 0.07} & \textbf{0.64 $\pm$ 0.08} & \textbf{0.87 $\pm$ 0.01}\\
\midrule
\multicolumn{5}{c}{\bfseries ALFWorld Dataset}\\
\midrule
Act-Only            & 1.95$\pm$0.05 & 0.65$\pm$0.02 & 0.51$\pm$0.01 & 0.62$\pm$0.14\\
ReAct               & 1.89$\pm$0.05 & 0.58$\pm$0.02 & 0.52$\pm$0.02 & 0.58$\pm$0.07\\
C-o-T               & 0.72$\pm$0.02 & 0.25$\pm$0.01 & 0.21$\pm$0.01 & 0.33$\pm$0.07\\
SPRING              & 1.59$\pm$0.03 & 0.54$\pm$0.02 & 0.47$\pm$0.01 & 0.58$\pm$0.03\\
\methodName{} (ours)& \textbf{2.31$\pm$0.06} & \textbf{0.74$\pm$0.02} & \textbf{0.62$\pm$0.02} & \textbf{0.69$\pm$0.01}\\
\midrule
\multicolumn{5}{c}{\bfseries ScienceWorld Dataset}\\
\midrule
Act-Only            & $2.05\pm0.08$ & $0.36\pm0.01$ & \textbf{0.18 $\pm$ 0.01} & $0.62\pm0.03$\\
ReAct               & $1.80\pm0.09$ & $0.31\pm0.02$ & $0.11\pm0.03$ & $0.44\pm0.04$\\
C-o-T               & $1.51\pm0.19$ & $0.26\pm0.03$ & $0.12\pm0.02$ & $0.63\pm0.03$\\
SPRING              & $1.90\pm0.38$ & $0.36\pm0.07$ & $0.11\pm0.07$ & $0.22\pm0.05$\\
\methodName{} (ours)& \textbf{2.49 $\pm$ 0.07} & \textbf{0.44 $\pm$ 0.01 } & $0.16\pm0.01$ & \textbf{0.74 $\pm$ 0.02}\\
\midrule
\multicolumn{5}{c}{\bfseries Business Process Dataset}\\
\midrule
Act-Only            & 3.40 $\pm$ 0.21 & 0.49 $\pm$ 0.03 & 0.37 $\pm$ 0.07 & 0.71 $\pm$ 0.07\\
ReAct               & 3.15 $\pm$ 0.24 & 0.52 $\pm$ 0.05 & \textbf{0.47 $\pm$ 0.07} & 0.60 $\pm$ 0.06\\
C-o-T               & 3.42 $\pm$ 0.24 & 0.52 $\pm$ 0.05 & 0.40 $\pm$ 0.07 & \textbf{0.74 $\pm$ 0.06}\\
SPRING              & 3.22 $\pm$ 0.22 & 0.47 $\pm$ 0.05 & 0.40 $\pm$ 0.07 & 0.65 $\pm$ 0.07\\
\methodName{} (ours)& \textbf{5.15 $\pm$ 0.07} & \textbf{0.71 $\pm$ 0.03} & 0.30 $\pm$ 0.06 & \textbf{0.74 $\pm$ 0.01}\\
\bottomrule
\end{tabular}
\end{table}

In \textbf{ScienceWorld}, \textbf{ALFWorld}, and \textbf{Business Process} datasets, FM reflects whether a task is successfully completed. However, FM alone can be misleading, since different methods may reach the same final action through divergent or error-prone trajectories. To assess procedural reliability, FM must be considered together with trajectory-based metrics (PML, PA, SM), which capture how consistently an agent stays aligned with the intended procedure. Furthermore, there exists more than one correct sequence of actions that can complete a task, it is possible for a method to achieve high FM, while having lower match scores PML, PA, and SM, which compare against a single reference trajectory. Ideally, all these metrics should be examined together to provide a more complete picture of performance.

On the \textbf{Business Process} dataset, \methodName{} achieves PML of $5.15\pm0.07$ beating the best baseline (C-o-T's $3.42\pm0.24$) and PA is $0.71\pm0.03$ compared to ReAct and C-o-T's PA score of $0.52\pm0.05$, while SM is $0.30\pm0.06$ because execution order is this dataset is flexible. \methodName{} attains an FM of \(0.74\pm0.01\). We observe that C-o-T achieves an FM score ($0.74\pm0.06$) close to \methodName{}, but its lower PML, PA, and SM indicate that it is less consistent in following the intended trajectory. This suggests that C-o-T can sometimes reach the correct final action, but with less stable alignment to the procedure, whereas \methodName{} maintains stronger trajectory fidelity.
On the \textbf{ALFWorld} dataset, \methodName{} consistently outperforms the baselines across all metrics, achieving a PML of $2.31\pm0.06$, PA of $0.74\pm0.02$, SM of $0.62\pm0.02$, and FM of $0.69\pm0.01$. 
On the \textbf{ScienceWorld} dataset, \methodName{} achieves a PML of $2.49\pm0.07$ against the best baseline’s $2.05\pm0.08$, and its plans match the reference more often (PA is $0.44\pm0.01$ vs. Act-Only's $0.36\pm0.01$). The SM score is slightly lower $0.16\pm0.01$ vs. $0.18\pm0.01$ for Act-Only, which is acceptable because many lab tasks let steps be taken in flexible order. Considering all metrics together, \methodName{} delivers more reliable performance overall.


On deeper inspection, we observe distinct shortcomings across different baselines. ReAct falls into reasoning–action loops (e.g., sensor $\rightarrow$ capture $\rightarrow$ sensor $\rightarrow$ capture), while C-o-T is prone to hallucinations when sketching long plans without grounding. Surprisingly, Act-Only occasionally performs at par with or even better than other baselines; we attribute this to its simplicity, which avoids compounding reasoning errors while navigating through complex procedures, though it lacks robustness across domains. In contrast, \methodName{}’s consistently stronger results across all datasets suggest that the explicit decision graph built during the \textit{Teach} phase yields a steadier plan, limiting early drift and cascading errors during execution. Because this graph is pre-computed, the \textit{Execute} phase is streamlined. The \emph{Execute} phase follows the graph, executes the prescribed actions and delivers the most stable performance across domains.


To analyze the impact of ambiguity, we specifically report results on the \emph{hard} category Recipe tasks in Table~\ref{tbl:hard-recipe}, where the tasks provide only a set of ingredients and the execution agent must infer the intended dish based on the provided ingredients before proceeding with the execution. \methodName{} delivers a clear margin over all baselines with FM reaching $0.81$, nearly triple that of C-o-T and SPRING, while SM climbing to $0.47$. These results highlight that while baselines struggle under ambiguity, the structured decision graph abstraction in \methodName{} enables more stable execution. Since real-world procedural tasks are inherently ambiguous, methods that can accommodate such uncertainty are essential. Notably, all baselines have visibility of the tool library from the outset, whereas \methodName{} first constructs the decision graph purely from procedure text, attaching tools to nodes only afterward based on the graph’s logical structure.





\begin{table}[t]
\centering
\caption{Performance on \emph{Hard} Category Recipe tasks (summarized over 5 independent trials). \methodName{} outperforms all baselines, showing strong robustness under ambiguity.}
\label{tbl:hard-recipe}
\begin{tabular}{@{}lcccc@{}}
\toprule
&\multicolumn{4}{c}{Metric}\\
\cmidrule(l){2-5}
\textbf{Method} & \textbf{PML} & \textbf{PA} & \textbf{SM} & \textbf{FM} \\
\midrule
Act-Only & $0.95 \pm 0.10$ & $0.24 \pm 0.02$ & $0.08 \pm 0.03$ & $0.28 \pm 0.04$ \\
CoT & $0.47 \pm 0.05$ & $0.11 \pm 0.01$ & $0.03 \pm 0.03$ & $0.12 \pm 0.04$ \\
ReAct & $0.69 \pm 0.06$ & $0.16 \pm 0.01$ & $0.05 \pm 0.00$ & $0.28 \pm 0.03$ \\
SPRING & $0.96 \pm 0.04$ & $0.21 \pm 0.02$ & $0.03 \pm 0.03$ & $0.18 \pm 0.07$ \\
\methodName{} (ours) & $\mathbf{2.27 \pm 0.32}$ & $\mathbf{0.54 \pm 0.08}$ & $\mathbf{0.47 \pm 0.07}$ & $\mathbf{0.81 \pm 0.04}$ \\
\bottomrule
\end{tabular}
\end{table}

\section{Related Work}
\label{sec.relatedwork}
Early LLM agents explored planning for embodied tasks and short horizons \cite{huang2022language}. Subsequent works interleaves reasoning and acting (ReAct) \cite{yao2023react} or chain-of-though planning before execution (C-o-T) \cite{wei2022chain}. In contrast, \methodName{} automatically produces an executable decision graph from free-form procedural text in a \textit{Teach} phase and leverages it during \textit{Execute} for context-aware execution.

LLMs have recently been applied to business procedures and workflows.  Kulkarni et al.\ \cite{kulkarni2025agen} drive SOP execution by selecting the next action label either an API call or a user prompt from a manually curated Global Action Repository. LLM4Workflow \cite{xu2024llm4workflow} embeds API schemas in the prompt and has an LLM emit a static XML workflow, while SmartFlow \cite{jain2024smartflow} converts on-screen GUI elements into text and lets an LLM generate a navigation script. ProcBench \cite{fujisawa2024procbench} introduces comprehensive metrics to evaluate LLM's multi-step procedure adherence, which we adopt in our evaluation. These methods follow a fixed execution path, in contrast, \methodName{} produces a decision graph that enables real-time branching and context-aware replanning.  

While above works focus on procedures, there is also a line of research on action execution in less formalized environments. SPRING \cite{wu2024spring} utilizes LLMs for reasoning-based decision-making in game environments (such as Crafter) which are not exactly business operating procedures but still relevant. SPRING employs a fixed DAG where nodes represent game-related questions and edges define their dependencies. DAG is traversed in topological order, computing LLM answers for each node. The final node poses the question of the best action to take, and the LLM's answer is directly translated into an executable action within the game. SPRING requires manual crafting of environment-related questions, which can be labor-intensive and limits scalability. AgentKit \cite{wu2024agentkit} extends SPRING by introducing dynamic components, allowing for modifications to the DAG at inference time by adding or removing nodes and dependencies.  However, it still relies on users to manually break down tasks into subtasks (nodes) of the DAG. In contrast, our methodology automatically creates graph-based representations, significantly reducing the manual effort required in SPRING and AgentKit approaches.



Finally, several works explore LLM-based code generation for task automation, generating complete solutions from text prompts or pseudo-code \cite{ni2024tree, wang2024executable, xu2024aios}. While powerful, these approaches do not model long-horizon procedures. By introducing an explicit decision-graph abstraction, \methodName{} enables dynamic execution across diverse domains.

\section{Conclusion}
\methodName{} introduces a two-phase framework: \textit{Teach}, which converts free-form procedures into decision graphs, and \textit{Execute}, which reuses these graphs across multiple task instances while adapting in real time to context. Unlike SPRING \cite{wu2024spring}, which requires hand-crafted question trees, or AgentKit \cite{wu2024agentkit}, which depends on manual task decomposition, \methodName{} generates decision graphs automatically, enabling greater scalability. Experiments across business, cooking, household, and science domains show state-of-the-art performance, demonstrating that structured representations combined with context-aware execution can handle diverse, long-horizon procedures and ambiguous tasks. Structured intermediate decision graph representations provide interpretability, support reuse and limit error accumulation through constrained execution paths. By separating procedure structuring from execution, \methodName{} ensures that once a procedure is structured, it becomes a reusable and reliable asset. We view \methodName{} as a step toward reliable, general-purpose procedure execution systems, automated where possible, but flexible enough to keep humans in the loop when domain sensitivity (e.g., in finance or healthcare) requires it.

\section{Limitations}
We acknowledge that our current evaluation relies primarily on GPT-4, which introduces cost considerations for large-scale deployment. However, the model-agnostic nature of our approach means that organizations can adapt \methodName{} to work with more cost-effective models, balancing performance requirements with operational constraints.

While \methodName{} is able to handle ambiguity in procedural tasks as shown in the \emph{hard} Recipe tasks (Table~\ref{tbl:hard-recipe}), its success still depends on the quality and completeness of the source procedural documentation. Our framework can handle various formats and styles of procedure text, however, it cannot compensate for fundamentally incomplete or incorrect source material.

\section{DISCLAIMER}
This paper was prepared for informational purposes by the Artificial Intelligence Research group of JPMorgan Chase \& Co and its affiliates (“J.P. Morgan”) and is not a product of the Research Department of J.P. Morgan. J.P. Morgan makes no representation and warranty whatsoever and disclaims all liability, for the completeness, accuracy or reliability of the information contained herein. This document is not intended as investment research or investment advice, or a recommendation, offer or solicitation for the purchase or sale of any security, financial instrument, financial product or service, or to be used in any way for evaluating the merits of participating in any transaction, and shall not constitute a solicitation under any jurisdiction or to any person, if such solicitation under such jurisdiction or to such person would be unlawful. © 2025 JPMorgan Chase \& Co. All rights reserved.

\bibliographystyle{ACM-Reference-Format} 
\bibliography{sample}


\end{document}